\documentclass{article}

\usepackage[preprint]{nips_2018}

\usepackage[utf8]{inputenc} 
\usepackage[T1]{fontenc}    
\usepackage{hyperref}       
\usepackage{url}            
\usepackage{booktabs}       
\usepackage{amsfonts}       
\usepackage{nicefrac}       
\usepackage{microtype}      
\usepackage[utf8]{inputenc}
\usepackage{amssymb}
\usepackage{fullpage}
\usepackage{xcolor}
\usepackage{amsmath,bm}
\usepackage{amsthm}
\usepackage{algorithmic}
\usepackage{algorithm}
\usepackage{graphicx}
\newcommand{\x}{\boldsymbol{x}}

\newcommand{\prob}{\mathbb{P}}
\newcommand{\N}{\mathcal{N}}
\usepackage{bbm}
\usepackage{caption}
\usepackage{subcaption}
\usepackage{hyperref}
\usepackage{color}

\title{Universal Decision-Based Black-Box Perturbations: Breaking Security-Through-Obscurity Defenses}

\author{
  Thomas A. Hogan\\
    Department of Mathematics\\
   University of California\\
  Davis, CA 95616 \\
  \texttt{tahogan@math.ucdavis.edu} \\
   \And
  Bhavya Kailkhura\\
 Center for Applied Scientific Computing\\
  Lawrence Livermore National Lab\\
  Livermore, CA 94550 \\
  \texttt{kailkhura1@llnl.gov} \\
}

\begin{document}

\maketitle

\begin{abstract}
We study the problem of finding a universal (image-agnostic) perturbation to fool machine learning (ML) classifiers (e.g., neural nets, decision tress) in the hard-label black-box setting.
Recent work in adversarial ML in the white-box setting (model parameters are known) has shown that many state-of-the-art image classifiers are vulnerable to universal adversarial perturbations: a fixed human-imperceptible perturbation that, when added to any image, causes it to be misclassified with high probability~\cite{wb1,wb2,wb3,param}. 
This paper considers a more practical and challenging problem of finding such universal perturbations in an obscure (or black-box) setting.
More specifically, we use zeroth order optimization algorithms to find such a universal adversarial perturbation when no model information is revealed-except that the attacker can make queries to probe the classifier. We further relax the assumption that the output of a query is continuous valued confidence scores for all the classes and consider the case where the output is a hard-label decision. 
Surprisingly, we found that even in these extremely obscure regimes, state-of-the-art ML classifiers can be fooled with a very high probability just by adding a single human-imperceptible image perturbation to any natural image.
The surprising existence of universal perturbations in a hard-label (or decision-based) black-box setting raises 
serious security concerns with the existence of a universal noise vector that adversaries can possibly exploit to
break a classifier on most natural images.

\end{abstract}

\begin{section}{Introduction}
Despite the remarkable success of machine learning approaches, especially deep neural networks (DNNs), in image classification, a series of results \cite{hardlabel,universal,three,four,five} has raised concerns regarding their utility in safety-critical applications. Attackers can design human-imperceptible perturbations which when added to almost any natural image result in their misclassification. This perturbation design problem is usually posed as an optimization problem: given a fixed classifier $f$ and a correctly classified image $\boldsymbol{x}_i$, find a noise vector $\epsilon_i$ of minimal norm, such that, $f(\boldsymbol{x}_i) \neq f(\boldsymbol{x}_i + \epsilon_i)$.  

There are a couple of issues with most of the current approaches developed to solve the above formulation. 
Most approaches assume a ``white-box" setting where the 
adversary has a complete knowledge of the ML model parameters. In such a setting, the gradient of the attack objective function can be computed
by back-propagation, and the perturbation design problem can be solved quite easily.
However, in many practical applications, model parameters (or internal design architecture) are inaccessible to the user~\cite{bb_social} and only queries can be made to obtain the corresponding outputs of the model (e.g., probability score, hard label (or top-$1$ predicted class), etc.).
This gives rise to Security through obscurity (STO) defense paradigm. STO is a process of implementing security within a system by enforcing secrecy and confidentiality of the system's internal design architecture. STO aims to secure a system by deliberately hiding or concealing its security flaws~\cite{sto}.
Interestingly, some recent works have shown that even in such a ``black-box" setting, it is possible to fool the ML classifier with a high probability~\cite{chen_zoo,papernot,bb_bhoj,sijia_nips,autozoom}. 
These black-box attacks can be broadly classified in two categories: $1)$ knowledge transfer based attacks, and, $2)$ zeroth-order optimization based attacks. In knowledge transfer based attacks, instead
of attacking the original model $f$, attackers try to construct a substitute model $\hat{f}$ to mimic $f$ and then attack it using existing white-box attack methods~\cite{papernot,bb_bhoj}. However, it was shown recently that these approaches usually leads to much larger distortion and low success rate of attack transfer~\cite{chen_zoo}. To overcome this limitation, zeroth-order optimization based attacks are devised which can be directly applied to minimize a suitable loss function for $f$ using derivative-free optimization or black-box optimization methods~\cite{chen_zoo,sijia_nips}.
In particular, \cite{chen_zoo} considered the problem of score-based black-box setting, where attackers can query the softmax layer output in addition to the final classification result. In this case, it is possible to reconstruct the original loss function and use a zeroth order optimization approach to optimize it. Most relevant to our work, the authors in~\cite{hlbbd, hardlabel} considered a  hard-label black-box setting which refers to cases where real-world ML systems only provide limited prediction results of an input query. Specifically, only the final decision (top-$1$ predicted label) instead of probability outputs is known to an attacker. 

However, compared to white-box attacks, all the black-box attacks discussed above are very computationally expensive (require millions of queries). Furthermore, these approaches are intrinsically dependent on the individual images: the perturbations are specifically crafted for each image independently. As a result, the computation of an adversarial perturbation for a new image requires solving an image-dependent optimization
problem from scratch, making their applicability in practice infeasible\footnote{There are some efforts on designing image-agnostic universal perturbation, however, are limited to the ``white-box" setting~\cite{universal,uni_gan,uni_white,uni_patch,uni_gen}. }.
This practical limitation gives us an impression that Security through obscurity (STO) defense may still be a viable solution. 
To evaluate this hypothesis, this paper considers the problem of finding a {\em Universal Adversarial Perturbation}- a vector $\epsilon$ which can be added to any image to fool the classifier with high probability, in a hard-label back-box setting. Such an $\epsilon$ would eliminate the need to recompute a new perturbation for each input. 
Note that the hard-label black-box setting is very challenging as it requires minimizing a non-continuous step function, which is combinatorial and cannot be solved by a gradient-based optimizer. The main contributions of this paper are as follows:
\begin{itemize}
    \item We show the existence of universal adversarial perturbations for ML classifiers in a hard-label black-box setting (breaking the STO defense).
    \item We reformulate the attack as an easy to solve continuous-valued optimization problem and propose a zeroth-order optimization algorithm for finding such perturbations.
    \item Experimental validations are performed on CIFAR10 dataset.
\end{itemize}


\section{Universal Hard-Label Black-Box Attacks}
We formalize in this section the notion of universal perturbations in a hard-label black-box setting,
and propose a method for finding such perturbations.
For simplicity, let us consider attacking a $K$-way multi-class classification model in this paper, i.e., $f:\mathbb{R}^d \rightarrow [K]$.
Finding a universal adversarial perturbation can be posed as a stochastic optimization problem: find $\epsilon$ of minimal norm such that 
\begin{equation}
\label{prob_uni}    \prob_{\x \in \N}(f(\x + \epsilon) \neq f(\x)) \geq p,
\end{equation}
where $\N$ is the set of natural images. The main focus of this
paper is to design approaches to find an image-agnostic minimal norm (or quasi-imperceptible) perturbation vector $\epsilon \in \mathbb{R}^d$
that fool the
classifier $f$ on almost all images sampled from $\mathcal{N}$ with only hard-labeled queries from $f$.
The main challenge now is to solve \eqref{prob_uni}
with only hard-labeled queries to $f$.

Note that the optimization problem \eqref{prob_uni} is extremely difficult to approach directly.
Not only is the gradient with respect to $\epsilon$ unavailable, but also the loss function ( indicator function of the set of $\epsilon$ satisfying equation \eqref{prob_uni}) is discontinuous. Furthermore, the loss function cannot be evaluated directly, and can only be estimated with empirical sampling. To overcome this challenge, we next introduce an auxiliary function which is much easier to optimize. Furthermore, we show that to optimize \eqref{prob_uni}, it suffices to optimize the auxiliary function.






\subsection{A Universal Auxiliary Function}
Now we  re-formulate universal hard-label black-box attack as another easier to optimize problem by defining a universal auxiliary objective function.
Later we will discuss
how to evaluate the function value using hard-label queries, and then apply a zeroth order optimization algorithm to obtain a universal adversarial perturbation. 

Let us consider the set of natural images $\mathcal{N}$ in the ambient space $\mathbb{R}^d$, where each $\bm{x} \in \mathcal{N}$ has true label $f(\bm{x}) \in [K]$, and the hard-label black-box function $f: \mathbb{R}^d \rightarrow [K]$. Following a similar approach as given in~\cite{hardlabel}, we define our new objective function\footnote{Note that under the manifold hypothesis $h$ is continuous almost everywhere, and is therefore amenable to zeroth-order optimization methods.} as:
\begin{equation}
\label{blackbox_obj}
    h(\theta,p) = argmin_{\lambda > 0}(\mathbb{P}_{\bm{x} \in \mathcal{N}}(f(\bm{x}+ \lambda \frac{\theta}{\|\theta\|}) \neq f(\bm{x})) \geq p).
\end{equation}
In this formulation, $\theta$ represents the universal search direction and $h(\theta,p)$ is the distance from $\mathcal{N}$ to the nearest universal adversarial perturbation along the direction $\theta$. In this formulation, instead of directly searching for a universal perturbation $\epsilon$, we search for the universal direction $\theta$ to
minimize the distortion $h(\theta,p)$, which leads to the following optimization problem:
\begin{equation}
  \theta^* := min_{\theta} h(\theta,p),  
\end{equation}
and in this case the universal perturbation vector and adversarial examples are given by
\begin{align}
    \epsilon = h(\theta^*,p) \frac{\theta^*}{\|\theta^* \|},\;\text{and}\;\; \bm{x}_i^* = \bm{x}_i+\epsilon.
\end{align}
Note that unlike white-box objective functions, which are discontinuous
step functions in the hard-label setting, $h(\theta,p)$ maps input direction to real-valued output (distance to decision boundary), which is usually a
continuous function. This makes it easy to apply
zeroth-order optimization methods to solve universal perturbation design problem.
In the absence of the knowledge of the image generating distribution, the most natural choice is to empirically estimate $h$ as follows:
take a sample set of $N$ images $\bm{x}_1, \dots,  \bm{x}_N$ and define
\begin{equation}
    \label{obj_fun}
    \hat{h}(\theta) = argmin_{\lambda > 0}(f(\bm{x}_i+ \lambda \frac{\theta}{\|\theta\|}) \neq f(\bm{x_i} ), \forall i \in [N]).
\end{equation}
In other words, $\hat{h}(\theta)$ is defined as the minimal perturbation distance in direction $\theta$ so that all $N$ images are misclassified.

\subsection{Algorithms to Find Universal Perturbation}
Even with our new auxiliary function $h$ defined, we cannot evaluate the gradient of $h$ due to the black-box nature of the problem. However, now we can evaluate the function values of $h$ (which are continuous) using the hard-label queries to the original classifier function $f$.
This procedure is used to find the initial $\theta_0$ and
corresponding $\hat{h}(\theta_0)$ in our optimization algorithm.  For a given normalized $\theta$, we do a fine-grained search and then a binary search similar to~\cite{hardlabel}. We omit the detailed algorithm for this part since it is similar to
Algorithm~\ref{evaluate2}.


\begin{algorithm}
\caption{Compute $\hat{h}(\theta)$ locally}
\label{evaluate2}
\begin{algorithmic}
\STATE {\bf Input:} Hard-label model $f$, original image label pairs $\{(\x_1, y_1), \dots (\x_N, y_N)\}$, query direction $\theta$, previous value $v$, increase/decrease ratio $\alpha = 0.01$, stopping tolerance $\epsilon$ (maximum tolerance of computed error)
 
\STATE $\theta \leftarrow \theta / \| \theta \|$
\IF {$f(\x_i + v\theta ) = y_i $ for some $i \in [N]$}
		\STATE {$v_{left} \leftarrow v$, $v_{right} \leftarrow (1+ \alpha )$}
        \WHILE {$f(\x_i+ v_{right} \theta ) = y_i $ for some $i \in [N]$} 
        	\STATE {$v_{right} \leftarrow (1+\alpha )v_{right}$}
        \ENDWHILE
\ELSE
	\STATE {$v_{right} \leftarrow v$, $v_{left} \leftarrow (1-\alpha)v $}
	\WHILE {$f(\x_i + v_{left} \theta )\neq y_i$ for all $i \in [N]$}
    	\STATE {$v_{left} \leftarrow (1-\alpha)v_{left}$}
    \ENDWHILE
    
\ENDIF
\STATE {\#\# Binary search within $[v_{left},v_{right}]$}
\WHILE {$v_{right} - v_{left} >\epsilon $}
	\STATE $v_{mid} \leftarrow (v_{right} + v_{left})/2$
    \IF {$f(\x_i + v_{mid}\theta) = y_i$ for some $i \in [N]$}
    	\STATE $v_{left} \rightarrow v_{mid}$
    \ELSE
    	\STATE $v_{right} \rightarrow v_{mid}$
    \ENDIF
\STATE {{\bf return} $v_{right}$}
\ENDWHILE

\end{algorithmic}
\end{algorithm}

\begin{algorithm}
\caption{RGF for universal hard-label black-box attack (See also \cite{hardlabel})}
\label{universalhardlabel}
\begin{algorithmic}
\STATE {\bf Input:} Hard-label model $f$, original image label pairs $\{(\x_1, y_1), \dots (\x_N, y_N)\}$, initial $\theta_0$.
\FOR {$t = 0, 1, 2, \dots, T$}
	\STATE {Randomly choose $\bm{u}_t$ from a zero-mean Gaussian distribution}
    \STATE {Evaluate $\hat{h}(\theta_t)$ and $\hat{h}(\theta_t + \beta \bm{u})$ using Algorithm~\ref{evaluate2}}
    \STATE {Compute $\hat{h'} = \frac{\hat{h}(\theta_t + \beta \bm{u}) - \hat{h}(\theta_t ) }{\beta} \bm{u}$}
    \STATE {$\theta_{t+1} \leftarrow \theta_t - \nu_t \hat{h'}$}
\ENDFOR
\STATE{ {\bf return} $x_0 + \hat{h}(\theta_T)\theta_T$}

\end{algorithmic}
\end{algorithm}

In practice, the optimization of $$\hat{h}(\theta) = argmin_{\lambda > 0}(f(\bm{x}_i+ \lambda \frac{\theta}{\|\theta\|}) \neq f(\bm{x_i} ), \forall i \in [N])$$
turns out to be difficult to minimize.  Therefore, in our implementation, we consider a couple of variants of $\hat{h}$ as discussed next next.
First, we consider a norm based approximation:
\begin{equation}
  \hat{h}_n(\theta) =  \| (\hat{h}(\theta ,\bm{x}_1), \dots, \hat{h}(\theta, \bm{x}_N))\|_{p}  
\end{equation}
where $\hat{h}(\theta ,\bm{x}_i)$ is obtained by applying $\hat{h}$ on a single image $i$ (instead of a set of images $\{1,\cdots,N\}$).
We call such an attack a ``NormAttack", as we are trying to minimize the $p$ norm of a vector with $h(\theta, \bm{x}_i)$ in each component.
Next, we consider a stochastic approximation
$\hat{h}_p$ as given below:
\begin{equation}
  \hat{h}_p(\theta,p) = argmin_{\lambda > 0}(f(\bm{x}_i+ \lambda \frac{\theta}{\|\theta\|}) \neq f(\bm{x_i} ), \text{ for at least } p\times N  \text{ of $i$ in $[N]$}.  
\end{equation}
We refer to this attack as the ``ProbAttack" as it aims to find the minimal perturbation needed to fool a proportion $p$ of the training samples.

Given $\hat{h}_n$ and $\hat{h}_p$, to solve the universal adversarial perturbation design problem for which we can only evaluate function value instead of gradient, zeroth-order optimization algorithms can be naturally applied. In this paper, we use Randomized Gradient-Free (RGF) method proposed in~\cite{rgf} as our zeroth-order algorithm. In each iteration, the gradient of function $h$ is estimated by
$$\hat{h'} = \frac{\hat{h}(\theta_t + \beta \bm{u}) - \hat{h}(\theta_t ) }{\beta} \bm{u}$$
where $\bm{u}$ is a random Gaussian vector, and $\beta>0$ is a smoothing parameter. 
\end{section}

\section{Experimental Results}
We test the performance of our universal hard-label black-box attack algorithms on convolutional neural network (CNN) models. 
We validate our approach on CIFAR-10 dataset. The network model used is as follows: four convolution layers, two max-pooling layers and two fully-connected layers. Using the parameters provided by~\cite{param}, we could achieve $82.5\%$ accuracy on CIFAR-10. 
All models are trained using Pytorch and our source code will be publicly available soon. 

\subsection{Breaking the STO Defense}
\begin{figure}[!t]
\begin{center}
\includegraphics[scale = .6]{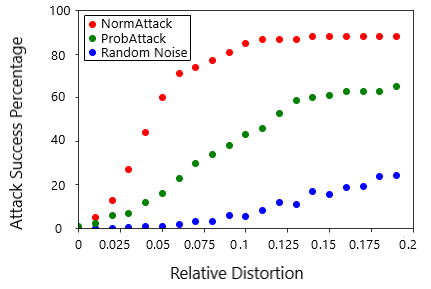}
\end{center}
\caption{Success rate of universal adversarial perturbations at various scales on $1000$ random images from CIFAR-10 dataset. The relative distortion is $L_2$ norm of perturbation divided by average norm of CIFAR-10 image. Since CIFAR-10 has only $10$ image classes, $90\%$ success rate is optimal for untargeted attack as it indicates that the classifier is essentially returning a random prediction.}\label{normsuccess}
\end{figure}

\begin{figure}[]
\includegraphics[scale = .95]{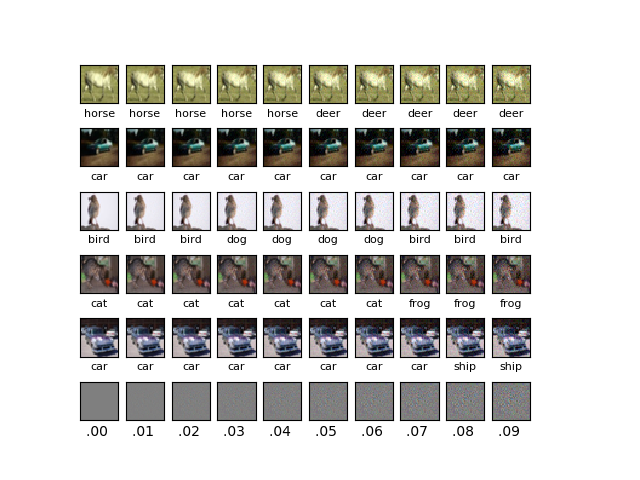}
\caption{Bottom row is perturbation scaled to various relative distortion levels. Other rows show perturbation applied to various images in CIFAR-10 images and predicted class.}\label{adv-images}
\end{figure}

In the first experiment, we analyze the robustness of deep neural network classifiers to black-box universal perturbations found using Algorithm~\ref{universalhardlabel} on CIFAR dataset. Specifically, we report the fooling ratio, that is the proportion of images that change labels when perturbed by our black-box universal perturbation. It can be seen from Fig.~\ref{normsuccess} that Algorithm~\ref{universalhardlabel} can find quasi-imperceptible perturbations which can fool DNN classifiers with a very high probability. Specifically, ``NormAttack" based black-box universal perturbation achieves high fooling/success rate with very small distortion. By increasing the magnitude of the distortion, we can achieve $100\%$ success rate as in ``white-box" attacks. 
We also show corresponding perturbed images for visual inspection of quasi-imperceptibility in Fig~\ref{adv-images}.  
These results show that even without accessing the model parameters, an adversary can fool DNN classifiers with only relying on hard-label queries to the black-box. As a consequence, STO based defenses are not robust for ML applications. 





\section{Conclusion}
In this paper, we showed the existence of hard-label black-box universal perturbations
that can fool state-of-the-art classifiers on natural images. 
We proposed an iterative algorithm to generate universal
perturbations without accessing the model parameters. 
In particular, we showed that these universal perturbations can be easily found using hard-labeled queries to ML black-box models, thereby, breaking the security through obscurity based defenses.  
Currently, we are devising techniques to utilize gradient information from white-box models (or knowledge transfer) to minimize the query-complexity of finding such hard-label black-box universal perturbations. Also, we plan to show that these universal perturbations generalize very well across different ML models resulting in doubly-universal perturbations (image-agnostic, network-agnostic).  
A theoretical analysis of existence of black-box universal perturbations
will be the subject of future research.

\section{Acknowledgements}
Thanks to NSF Mathematical Sciences Research Program for financial support for this research. 
Bhavya Kailkhura's work was performed under the auspices of  the U.S. Department of Energy by Lawrence Livermore National Laboratory under Contract DE-AC52-07NA27344 (LLNL-CONF-761205-DRAFT). Thomas Hogan would like to express gratitude to Lawrence Livermore National Lab, in particular to Bhavya Kailkhura and Ryan Goldhahn, for their hospitality during his internship where this research was conducted.

\bibliographystyle{plainnat}

\bibliography{references}

\end{document}